\documentclass[10pt,twocolumn,letterpaper]{article}

\usepackage[pagenumbers]{wacv}

\usepackage{graphicx}
\usepackage{amsmath}
\usepackage{amssymb}
\usepackage{booktabs}
\usepackage{algorithm}
\usepackage{marvosym}
\usepackage{algpseudocode}
\usepackage{xcolor}

\usepackage[pagebackref,breaklinks,colorlinks]{hyperref}
\usepackage{lipsum}  

\usepackage[capitalize]{cleveref}
\crefname{section}{Sec.}{Secs.}
\Crefname{section}{Section}{Sections}
\Crefname{table}{Table}{Tables}
\crefname{table}{Tab.}{Tabs.}

\begin{document}

\title{AWF: Adaptive Weight Fusion for Enhanced Class Incremental Semantic Segmentation}

\author{
Zechao Sun\textsuperscript{1*}\quad
Shuying Piao\textsuperscript{1*}\quad
Haolin Jin\textsuperscript{2}\quad
Chang Dong\textsuperscript{1}\\
Lin Yue\textsuperscript{1}\quad
Weitong Chen\textsuperscript{1$\dagger$} \quad
Luping Zhou\textsuperscript{2$\dagger$} \\
\textsuperscript{1}The University of Adelaide \quad
\textsuperscript{2}The University of Sydney 
}

\maketitle

\AddToShipoutPicture*{%
  \put(50,65){
    \parbox{\textwidth}{%
      \noindent\rule{0.4\textwidth}{0.2pt}\\[-0.0em]
      \footnotesize\textsuperscript{*}Equal contribution. \textsuperscript{†}Indicates the corresponding author.
    }%
  }%
}

\begin{abstract}
  Class Incremental Semantic Segmentation (CISS) aims to mitigate catastrophic forgetting by maintaining a balance between previously learned and newly introduced knowledge. Existing methods, primarily based on regularization techniques like knowledge distillation, help preserve old knowledge but often face challenges in effectively integrating new knowledge, resulting in limited overall improvement. Endpoints Weight Fusion (EWF) method, while simple, effectively addresses some of these limitations by dynamically fusing the model weights from previous steps with those from the current step, using a fusion parameter alpha determined by the relative number of previously known classes and newly introduced classes. However, the simplicity of the alpha calculation may limit its ability to fully capture the complexities of different task scenarios, potentially leading to suboptimal fusion outcomes. In this paper, we propose an enhanced approach called Adaptive Weight Fusion (AWF), which introduces an alternating training strategy for the fusion parameter, allowing for more flexible and adaptive weight integration. AWF achieves superior performance by better balancing the retention of old knowledge with the learning of new classes, significantly improving results on benchmark CISS tasks compared to the original EWF. And our experiment code will be released on Github.
\end{abstract}

\section{Introduction}
\label{sec:intro}

Semantic segmentation is a key task in various visual applications, including object recognition\cite{pathak2018application}, medical imaging\cite{jiang2018medical}, and autonomous driving \cite{grigorescu2020survey}. Traditional fully-supervised approaches focus on segmenting a fixed set of classes predefined in the training phase. However, as real-world applications evolve, models need to incrementally learn new classes without forgetting previously acquired knowledge. A naive solution is to retrain the model using a combination of old and new data, but this approach is both computationally expensive and requires extensive manual 
\begin{figure*}
  \centering
  \begin{subfigure}{0.52\linewidth}
     \includegraphics[height=2.55in, width=\linewidth]{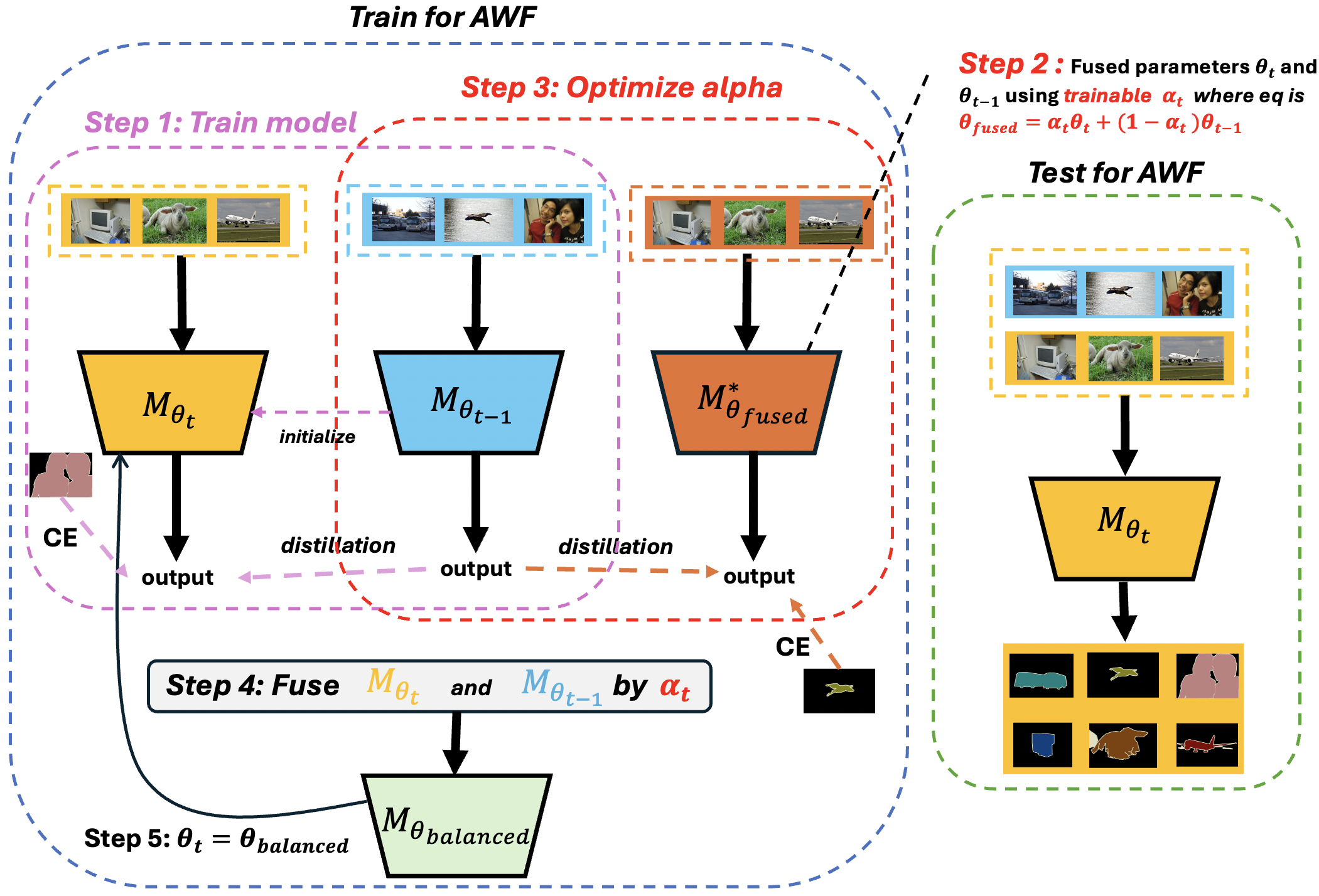}
    \caption{A single cycle\textbf{(step 1 to step 5)} of Adaptive Weight Fusion (AWF) training, This cycle will repeat over multiple times until training is complete.}
    \label{fig:1a}
  \end{subfigure}
  \hfill
  \begin{subfigure}{0.47\linewidth}
    \includegraphics[height=2.55in,width=\linewidth]{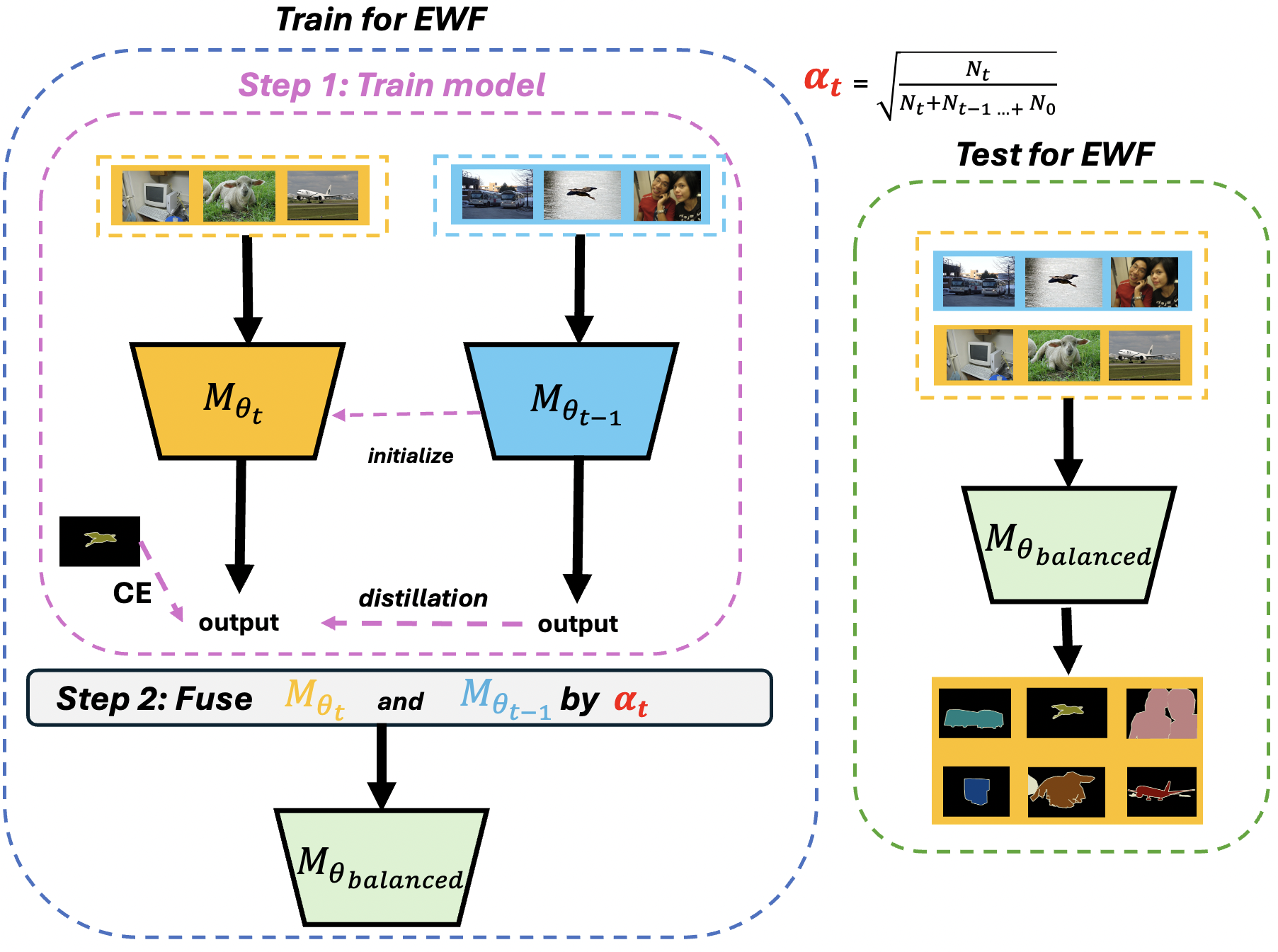}
    \caption{EWF method with entire training \textbf{(step 1 to step 2)} process without alternating steps.}
    \label{fig:1b}
  \end{subfigure}
  \caption{Illustration of the Endpoints Weight Fusion (EWF) process and our Adaptive Weight Fusion process (AWF). \(M_{\theta_t}\) represent model for current step, \(M_{\theta_{t-1}}\) represent model for last step, \(M_{\theta_{fused}}\) represent a new branch with the fused parameters \(\theta_{fused}\) for optimize trainable $\alpha$ during alternative training. $N_{t}$ refers to the number of new added classes in the task t.}
  \label{fig:1}
\end{figure*}
labeling. Alternatively, fine-tuning the model on new data can lead to overfitting and rapid forgetting of old classes, a phenomenon known as catastrophic forgetting \cite{mccloskey1989catastrophic}.

Class Incremental Semantic Segmentation (CISS) \cite{Cermelli_2020_CVPR,Douillard_2021_CVPR,zhang2022representation} has been introduced to mitigate catastrophic forgetting, with the goal of balancing the preservation of old knowledge and the acquisition of new knowledge, without the need to access to old training data. This is particularly important in scenarios where data privacy regulations or storage limitations prevent the reuse of previously collected data. Many CISS methods \cite{Cermelli_2020_CVPR,Douillard_2021_CVPR,douillard2020podnet,michieli2021continual} rely on regularization methods, which encourages the model to retain knowledge about old classes by imposing regularization constraints during training. Although knowledge distillation has been effective in alleviating forgetting, it still struggles when the old classes in the newly added data are incorrectly labeled as background. This mislabeling amplifies overfitting to the new classes and results in poor segmentation performance for previously learned classes.

In addition to regularization-based methods, another class of solutions has focused on model fusion strategies, where the knowledge of an old model is fused with that of a newly trained model. These methods \cite{he2019rethinking,Liu_2021_CVPR,singh2021rectification,Yan_2021_CVPR} often involve expanding the model with additional parameters or ensembling multiple models, which increases computational complexity and inference time. Compression-based techniques \cite{Yan_2021_CVPR,wang2022foster} attempt to reduce model size but often lead to a bias towards new data, as old knowledge may be underrepresented in the fusion process. Some approaches \cite{Zhu_2022_CVPR,zhang2022representation} rely on reparameterization, which fuse model components at the parameter level. While effective, they are often constrained by the need for architecture-specific operations. In response to these limitations, the Endpoints Weight Fusion (EWF) method \cite{xiao2023endpoints} was introduced, which integrates regularization techniques \cite{Cermelli_2020_CVPR,Douillard_2021_CVPR} and fuses the weights of the old and new models using a simple yet effective dynamic factor alpha, determined by the ratio of the relative number of previously known classes and newly introduced classes(as shown in Figure 1b). EWF offers a significant advantage by avoiding further training and maintaining a constant model size. However, the simplicity of EWF’s weight fusion mechanism can sometimes lead to suboptimal results, particularly in complex scenarios where the fixed alpha fails to fully capture the relationship between old and new knowledge.

In this work, we propose an improved method, Adaptive Weight Fusion (AWF), to address the limitations of EWF in Class Incremental Semantic Segmentation. In addition to incorporating typical knowledge distillation methods \cite{Cermelli_2020_CVPR,Douillard_2021_CVPR} that have proven to enhance the performance of model fusion techniques. AWF introduces a dynamic and trainable fusion parameter alpha, which is optimized through alternating training epochs(as shown in Figure \ref{fig:1a}). By introducing alternating training earlier in the process, the fusion dynamically adapts to the changing data characteristics throughout the training. This helps prevent the model from overly focusing on the current task's data, ensuring a more balanced integration of new and old knowledge without access to previous data.

To summarize, the main contributions of this paper are:

\begin{itemize}
    \item We propose an Adaptive Weight Fusion (AWF) strategy, which maintains the same model size and introduces a more dynamic and trainable fusion parameter optimized through alternating training. AWF effectively balances old and new knowledge, which alleviates catastrophic forgetting more efficiently compared to EWF\cite{xiao2023endpoints}.
    
    \item Our method can be seamlessly integrated with several typical Knowledge distillation methods,In most of CISS tasks, our AWF method consistently improves upon the baseline EWF\cite{xiao2023endpoints} by more than 1\%.
    
    \item We conduct extensive experiments on several CISS benchmarks, including PASCAL VOC and ADE20K, demonstrating that AWF significantly outperforms baseline methods, achieving a state-of-the-art performance across various scenarios.
    
\end{itemize}

\section{Related Work}
\label{sec:related_work}
\subsection{Class Incremental Learning}
\label{subsec:CIL}
\begin{figure*}[t]
  \centering
   \includegraphics[height=2.7in,width=0.97\textwidth]{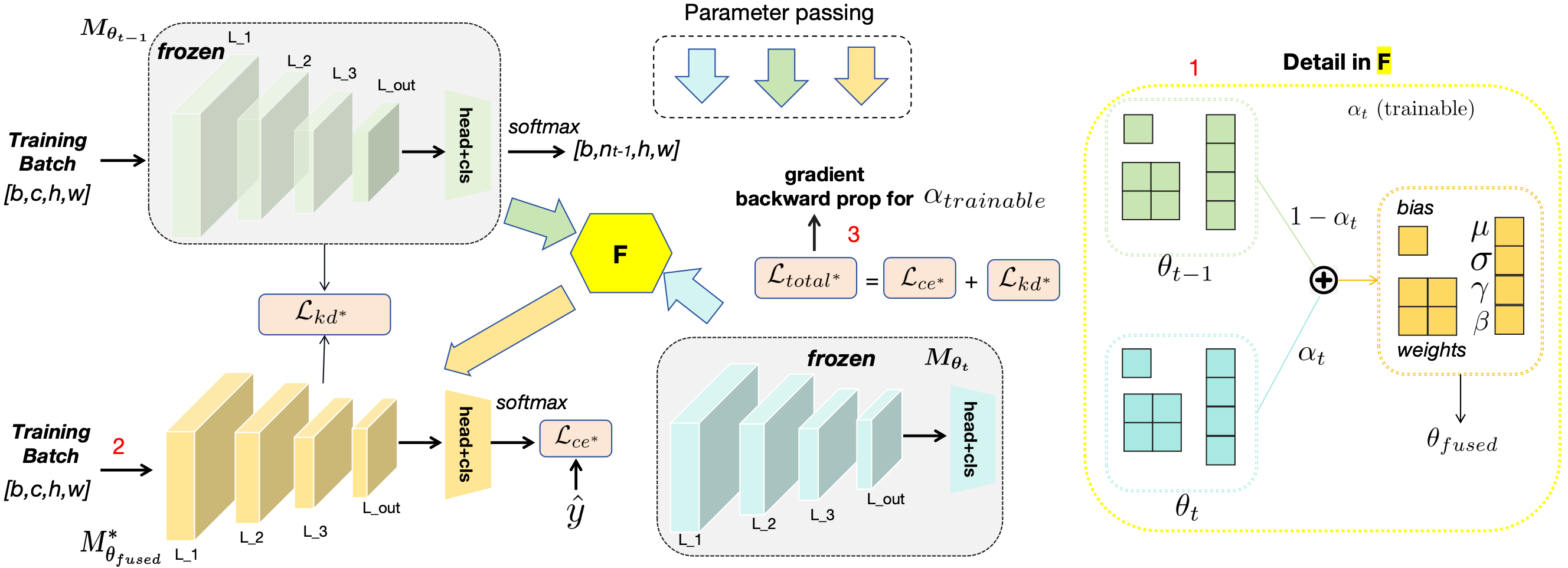}
   \caption{Illustration of an alpha training process(step \textcolor{red}{1} to \textcolor{red}{3}). In the lower left corner. We illustrated the new branch with fused parameters \(M^*_{\theta_{fused}}\), which is use for training alpha with no further computation cost, because we freeze \(M_{\theta_{t}}\) before start alpha training.}
   \label{fig:onecol}
\end{figure*}
Class Incremental Learning (CIL) primarily focuses on mitigating \emph{catastrophic forgetting} while enabling models to learn new classes incrementally. This challenge arises when models overwrite old knowledge with new knowledge. Existing CIL methods can be categorized into three main approaches \cite{de2021continual}. \emph{Replay-based methods} \cite{bang2021rainbow,Belouadah_2019_ICCV,Kim_2021_ICCV} store a small subset of old data and periodically mix it with new data to re-train the model in order to maintain performance on prior tasks. Although effective, this approach can introduce memory overhead and also raise privacy concerns . \emph{Regularization Based Method} \cite{Chaudhry_2018_ECCV,chaudhry2021using,Dhar_2019_CVPR,douillard2020podnet}
or \emph{Knowledge distillation techniques} \cite{Rebuffi_2017_CVPR,Simon_2021_CVPR}, on the other hand, aim to retain old knowledge without storing data, using intermediate representations or soft targets. These methods reduce \emph{memory consumption} but can increase \emph{computational costs} and may constrain the model's ability to fully learn new knowledge, as the focus remains on preserving prior information. Lastly, some works focus on \emph{Structural Based Method} \cite{NEURIPS2020_b3b43aee,Smith_2021_ICCV,Verma_2021_CVPR}, which freeze old models and
expand the architecture space to learn new knowledge, though they often lead to larger \emph{model sizes} as tasks increase. Overall, CIL approaches seek to balance the retention of old knowledge with the flexibility to learn new knowledge.

\subsection{Class Incremental Semantic Segmentation}
\label{subsec:CISS}
Class Incremental Semantic Segmentation (CISS)\cite{Cermelli_2020_CVPR,Douillard_2021_CVPR} is an extension of the continual learning paradigm focused on the task of semantic segmentation\cite{garcia2017review}. In CISS, models must continually learn to assign labels to each pixel of an image while maintaining knowledge of previously learned classes. Unlike image classification, semantic segmentation involves dense prediction at the pixel level, making it more memory-intensive and computationally demanding. So various Distillation-based methods was proposed and have become popular as they transfer old knowledge to new models without retaining data from previous tasks. For example, MiB\cite{Cermelli_2020_CVPR} uses a modeling strategy to account for potential class shifts and applies logits distillation to constrain representation ability, while PLOP\cite{Douillard_2021_CVPR} use pseudo labeling method to mitigate background shift and applies feature distillation to constrain representation ability. Another approach, like SSUL\cite{NEURIPS2021_5a9542c7}, avoids distillation altogether by fixing the feature extractor, yet this can disrupt the balance between plasticity and stability. SDR\cite{Michieli_2021_CVPR} uses prototype matching to enhance consistency in the latent space by ensuring that features learned from new data align with those learned from previous tasks, RC-IL\cite{zhang2022representation} addresses the limitations of strip pooling by introducing an average-pooling-based distillation mechanism.
\subsection{Weight Fusion Methods}
\label{subsec:WF}

Weight fusion is widely used in neural network training to enhance model performance by combining weights from different sources. In the linear mode, approaches like ACNet\cite{Ding_2019_ICCV} and RepVGG\cite{Ding_2021_CVPR} use structural re-parameterization to merge multi-branch layers into a single Convolution laye. In the nonlinear mode, weight averaging, as seen in methods like BYOL\cite{Kornblith_2019_CVPR}, uses techniques such as Exponential Moving Averages(EMA)\cite{article} to improve knowledge transfer and model stability across different tasks. In the context of continual learning, EWF\cite{xiao2023endpoints} focus on combining old and new model parameters to achieve a balance between old and new knowledge. EWF fuses old and new model weights based on a dynamic factor without requiring additional training, making it efficient in both memory and computational cost. However, it may struggle in scenarios where a relatively fixed fusion factor cannot fully capture the relationships between old and new knowledge. Our proposed Adaptive Weight Fusion (AWF) addresses these limitations by introducing a trainable fusion parameter that is optimized during the training process.

\section{Method}
\label{sec:Method}
\subsection{Preliminaries}
We adopt a multi-stage training framework where the model \( M_{\theta} \) learns sequentially over \( T \) tasks in a fully supervised semantic segmentation setting. The model state after completing task \( t \) is represented as \( M_{\theta_t} \). For each task \( t \), the dataset \( D^t = \{x_i, y_i\} \) contains input data \( x_i \in \mathbb{R}^{C \times H \times W} \) and corresponding ground truth labels \( y_i \in \mathbb{R}^{H \times W} \). The label space for each task is given by \( C^t \cup \{c_b\} \), where \( C^t \) represents the set of classes introduced in the current task and \( c_b \) is the background class. Since each step introduces entirely new classes, there is no overlap between the sets of classes from different tasks. As a result, the model must handle disjoint class distributions across steps, which often leads to catastrophic forgetting. Furthermore, to reduce the burden of annotation, only the current task’s categories are labeled, causing \( c_b \) to represent not only the actual background but also classes from previous and future tasks. This variation in the meaning of \( c_b \) across tasks complicates the training process because \( M_{\theta_t} \) must distinguish between actual background pixels and those belonging to classes that were learned in previous tasks. This ambiguity increases the difficulty of retaining old knowledge, thereby heightening the risk of catastrophic forgetting. In addition, the fuser, denoted as \( F \), which optimizes the fusion parameter \( \alpha \) to merge knowledge from different tasks, and a secondary model branch \( {M_{\theta}^*} \), which follows the same behavior as the primary model \( M_{\theta} \), but is used specifically to optimize \( \alpha \) during the alpha training process.

\subsection{Adaptive weight fusion}
Endpoints Weight Fusion (EWF)\cite{xiao2023endpoints} has shown significant advantages in balancing old and new knowledge. The key idea behind EWF is to use one dynamic factor $\alpha_t$ for fusing the model parameters between \(M_{\theta_t}\) and \(M_{\theta_{t-1}}\) after completing task t. The Eq. \ref{eq:1} computes the fusion parameter \( \alpha_t \): 
\begin{equation}
  \alpha_t = \sqrt{\frac{N_{t}}{N_{t} + N_{t-1} + N_{t-2} + \cdots + N_{0}}}
  \label{eq:1}
\end{equation}
where \( N_{t} \) represents the number of new classes to be added in the incremental step t.
The Eq.\ref{eq:2} shows using \( \alpha_t \) to balance  the parameters of the model from the previous task \( \theta_{t-1} \) with the parameters learned from the current task \( \theta_t \):
\begin{equation}
  \theta_{\text{balanced}} = \alpha_t \theta_t + (1 - \alpha_t) \theta_{t-1}
  \label{eq:2}
\end{equation}
After the model finishes training on the current task, these two formulas are used to fuse the old and new parameters. Although \( \alpha_t \) is dynamic, it is not adaptive enough to fully capture the evolving relationship between old and new knowledge. Therefore, there is still room for improvement in making the fusion process more flexible and responsive to the current task's demands.
At each step \( t \), we also have two key model states: \( \theta_{t-1} \) and \( \theta^i_t \ \). In the AWF approach, these two models are fused dynamically using a learnable parameter \( \alpha_{trainable} \), optimized within a fusion mechanism denoted as \( F(\theta_{t-1}, \theta^i_t) \).
The fusion operation can be described as:

\begin{align}
\theta^i_{\text{fused}} &=  F(\theta_{t-1}, \theta^i_t) \nonumber\\
&= \alpha^i_{\text{trainable}}  \theta^i_t + (1 - \alpha^i_{\text{trainable}})  \theta_{t-1}
\label{eq:3}
\end{align}

Here, \( \theta^i_{\text{fused}} \) represents the fused parameters at iteration $i$, which corresponds to a specific time point during the alpha training process. while \( \alpha^i_{trainable} \) is a trainable parameter at iteration $i$ that dynamically adjusts the balance between retaining old knowledge and learning new classes. 

\textbf{Knowledge distillation and EWF alpha initialization method are enhanced for AWF}. During alternating model training, we utilize knowledge distillation to constrain the output and feature representations between \( M_{\theta_{t-1}} \) and \( M_{\theta_t} \). We apply two forms of distillation: feature-based distillation\cite{Douillard_2021_CVPR} and logit-based distillation\cite{Cermelli_2020_CVPR}, with their respective losses defined as follows:
\begin{equation}
L_{FKD} = \frac{1}{|D|} \sum_{(x_i, y_i) \in D} \left\| \Psi_{t-1}(x_i) - \Psi_{t}(x_i) \right\|^2
\label{eq:4}
\end{equation}
\begin{equation}
L_{LKD} = \frac{1}{|D|} \sum_{(x_i, y_i) \in D} KL\left( \Phi_{t-1}(\Psi_{t-1}(x_i)), \Phi_{t}(\Psi_{t}(x_i)) \right)
\label{eq:5}
\end{equation}
where \( \Psi \) represents the feature embeddings, and \( \Phi \) denotes the output logits. Our alternating training method is divided into two phases: model training and alpha training. In the model training phase, we optimize the \(M_{\theta_t}\) on the task t using Cross-entropy loss and apply knowledge distillation loss(Eq.\ref{eq:4}, Eq.\ref{eq:5}) to ensure the \(M_{\theta_t}\) retains important knowledge from previous \(M_{\theta_{t-1}}\) while adapting to new information. Figure.\ref{fig:onecol} shows one iterations of alpha training. Before we start alpha training, we freeze parameters \(\theta_t\) in model \(M_{\theta_{t}}\) and initialize the trainable fusion parameter \( \alpha_{trainable} \) using the Eq.\ref{eq:1}. Since EWF has already demonstrated strong performance in balancing old and new knowledge using dynamic \(\alpha\) that calculate by Eq.\ref{eq:1},  using Eq.\ref{eq:1} to initialize \( \alpha_{trainable} \) in AWF allows us to shorten the training process for \( \alpha_{trainable} \), leading to faster convergence and improved overall optimization. During alternating alpha training, The objective of AWF is to optimize the fusion parameter \( \alpha_{trainable} \), To achieve this, we minimize a combined loss function that consists of a knowledge distillation loss \( L_{\text{FKD}^*} \) or \(L_{\text{LKD}^*}\) and a cross-entropy loss \( L_{\text{CE}^*} \).  They can be represented as:
\begin{equation}
L_{{FKD}^*} = \frac{1}{|D|} \sum_{(x_i, y_i) \in D} \left\| \Phi_{\text{fused}}(x_i) - \Phi_{t-1}(x_i) \right\|^2
\end{equation}
\begin{equation}
L_{{LKD}^*} = \frac{1}{|D|} \sum_{(x_i, y_i) \in D} KL\left( \Phi_{t-1}(\Psi_{t-1}(x_i)), \Phi_{\text{fused}}(\Psi_{\text{fused}}(x_i)) \right)
\end{equation}
\begin{equation}
L_{{CE}^*} = \frac{1}{|D|} \sum_{(x_i, y_i) \in D} {CE}\left(M^*_{\theta_{\text{fused}}}(x_i), \hat{y}_i \right)
\end{equation}
where \( \hat{y}_i \) represents the pseudo labels, and the cross-entropy loss \( CE \) is defined as:
\begin{equation}
{CE}(M^*_{\theta_{\text{fused}}}(x_i), \hat{y}_i) = - \sum_{c} \hat{y}_i^c \log \left( p(M^*_{\theta_{\text{fused}}}(x_i) = c) \right)
\end{equation}
Since we cannot access \{{\(D^{0}\), \(D^{1}\)... \(D^{t-1}\)}\}, training \(M_{\theta_t}\) for too long before switching to alpha training phase would cause the \(M_{\theta_t}\) to fit too well to the \(D^{t}\). this would cause \(\alpha_{trainable}\) to be heavily biased towards the \(\theta_t\) during alpha training phase, rather than finding a more better balance between the \(\theta_{t-1}\) and \(\theta_t\). To address this, I train the \(M_{\theta_t}\) for only a few epochs before switching to optimize alpha training phase. For EWF\cite{xiao2023endpoints}, they train \(M_{\theta_t}\) for N epochs before fusion \(M_{\theta_t}\) with \(M_{\theta_{t-1}}\) by \(\alpha\), our AWF just train \(M_{\theta_t}\) for N/3 epochs before switching to optimize alpha training phase, after alpha training phase, then train \(M_{\theta_t}\)..., repeating this process until the total number of training epochs for \(M_{\theta_t}\) reaches N. 
\begin{algorithm}
    
    \caption{Pseudo code for AWF in incremental steps}
    \footnotesize
    \label{alg:A1}
    \begin{algorithmic}[1]
    
        \Require model $M_{\theta}$, new branch to train alpha $M^*_{\theta}$, initial parameters $\theta_0$, num of tasks $T$, dataset $D_T$, learning rates for model $\gamma_\theta$ and for alpha $\gamma_\alpha$, num of model training epochs $E_{\theta}$ and num of alpha training epochs $E_{\alpha}$, $\text{trainable}$ $\alpha$, Fuser $F$ 
    
        \State $t \gets 1$
        \While{$t \leq T$}
        \State Initialize $N_{\text{new}}, N_{\text{old}}$
        \State $\alpha_t \gets \sqrt{\frac{N_{\text{new}}}{N_{\text{new}} + N_{\text{old}}}}$
        \State $i \gets 1$
        \While{not converged}
            \State Sample mini-batch $\{x_i, y_i\} \sim D$
            
            \If{$(i-1)\bmod(E_{\theta} + E_{\alpha}) >= E_{\theta}$}  
            \State 
            $\theta^i{_{\text{fused}}} \gets F(\theta^i_t, \theta^1)$ \Comment{\textcolor{red}{\(\alpha\) training phase}}
            \State $y^t_i \gets M^*_{\theta^i_{\text{fused}}}(x_i)$
            \State $y^{t-1}_i \gets M^*_{\theta_{t-1}}(x_i)$
            \State $\nabla_{\alpha} L_{\text{total}^*} \gets \nabla_{\alpha} L_{\text{CE}^*}(y^t_i, \hat{y_i}) + \nabla_{\alpha} L_{\text{KD}^*}(y^t_i, y^{t-1}_i)$
            \State $\alpha_t^{i+1} \gets \alpha_t^i - \gamma_\alpha \nabla_{\alpha} L_{\text{total}^*}$
            \If{$(i-1)\bmod(E_{\theta} + E_{\alpha}) == (E_{\theta} + E_{\alpha} - 1)$}
            \State $\theta_{\text{balanced}} \gets \alpha_t^{i+1} \theta^i_{t} + (1 - \alpha_t^{i+1}) \theta_{t-1}$
            \State $\theta^i_t \gets \theta_{\text{balanced}}$
            \EndIf
            \Else
            \State $y^t_{i} \gets M_{\theta_t}(x_i)$  \Comment{\textcolor{red}{model training phase}}
            \State $\nabla_{\theta} L_{\text{total}} \gets \nabla_{\theta}L_{\text{CE}}(y^t_i, \hat{y}) + \nabla_{\theta}L_{\text{KD}}(y^t_{i}, M_{\theta_{t-1}}(x_i))$
            \State $\theta^{i+1}_t \gets \theta^i_t - \gamma_\theta\nabla_{\theta}L_{\text{total}}$
            
            \EndIf
            \State $i \gets i + 1$
        \EndWhile
        \State $t \gets t + 1$
        \EndWhile
    \end{algorithmic}
    \normalsize
\end{algorithm}
\subsection{Overall Framework}
Consider both \( L_{\text{LKD}} \) and \( L_{\text{FKD}} \) as forms of \( L_{\text{KD}} \). The total loss during alternating model training is defined as:
\begin{equation}
L_{{total}} = L_{{CE}} + L_{{KD}}
\end{equation}
Our overall objective for model training phase is to minimize the total loss by adjusting the model parameters \(\theta_t\) . which is updated via gradient descent as follows:
\begin{equation}
\theta_t \leftarrow \theta_t - \gamma_\theta \frac{\partial L_{{total}^*}}{\partial \theta_t}
\end{equation}
Consider both \( L_{\text{LKD}^*} \) and \( L_{\text{FKD}^*} \) as forms of \( L_{\text{KD}^*} \). The total loss during alternating alpha training is defined as:
\begin{equation}
L_{{total}^*} = L_{{CE}^*} + L_{{KD}^*}
\end{equation}
Our overall objective for alpha training phase is to minimize the total loss by adjusting the fusion parameter \( \alpha_{trainable} \). The parameter \( \alpha_{trainable} \) is updated via gradient descent as follows:
\begin{equation}
\alpha_{\text{trainable}} \leftarrow \alpha_{\text{trainable}} - \gamma_\alpha \frac{\partial L_{{total}^*}}{\partial \alpha_{\text{trainable}}}
\end{equation}
And the overall algorithm of AWF is shown in Alg.\ref{alg:A1}.

\section{Experiments}
In this part, we outline the experimental protocols, scenarios, and training details. Additionally, we provide a comprehensive evaluation of our algorithm through both quantitative and qualitative experiments.
\begin{table*}[t] 
  \centering
  \resizebox{\textwidth}{!}{
  \small
  \begin{tabular}{@{}l|ccc|ccc|ccc|ccc@{}}
    
    \toprule
     & \multicolumn{3}{c}{\textbf{15-1} (6 steps)} & \multicolumn{3}{c}{\textbf{10-1} (11 steps)} & \multicolumn{3}{c}{\textbf{5-3} (6 steps)} & \multicolumn{3}{c}{\textbf{19-1} (2 steps)} \\
 
    \textbf{Method} & 0-15 & 16-20 & all & 0-10 & 11-20 & all & 0-5 & 6-20 & all & 0-19 & 20 & all \\
    \midrule
    LwF\cite{8107520} (TPAMI2017) & 6.0 & 3.9 & 5.5 & 8.0 & 2.0 & 4.8 & 20.9 & 36.7 & 24.7 & 53.0 & 8.5 & 50.9 \\
    ILT\cite{Michieli_2019_ICCV} (ICCVW2019) & 9.6 & 7.8 & 9.2 & 7.2 & 3.7 & 5.5 & 22.5 & 31.7 & 29.0 & 68.2 & 12.3 & 65.5 \\
    SDR\cite{Michieli_2021_CVPR} (CVPR2021) & 47.3 & 14.7 & 39.5 & 32.4 & 17.1 & 25.1 & - & - & - & 69.1 & 32.6 & 67.4 \\
    RCIL\cite{zhang2022representation} (CVPR2022) & 70.6 & 23.7 & 59.4 & 55.4 & 15.1 & 34.3 & 63.1 & 34.6 & 42.8 & 77.0 & 31.5 & 74.7 \\
    \midrule
    MiB\cite{Cermelli_2020_CVPR} (CVPR2020) & 38.0 & 13.5 & 32.2 & 12.2 & 13.1 & 12.6 & 57.1 & 42.5 & 46.7 & 71.2 & 22.1 & 68.9 \\
    MiB + EWF\cite{xiao2023endpoints} (CVPR2023) & 78.0 & 25.5 & 65.5 & 56.0 & 16.7 & 37.3 & 69.0 & 45.0 & 51.8 & 77.8 & 12.2 & 74.7 \\
    MiB + \textbf{AWF(ours)} & 78.1 & 25.7 & 65.6 & 56.4 & 18.2 & 38.2 & 72.2 & 48.0 & \textbf{54.9} & 78.3 & 25.8 & \textbf{75.8} \\
    \midrule
    PLOP\cite{Douillard_2021_CVPR} (CVPR2021) & 65.1 & 21.1 & 54.6 & 44.0 & 15.5 & 30.5 & 25.7 & 30.0 & 28.7 & 75.4 & 37.3 & 73.5 \\
    PLOP + EWF\cite{xiao2023endpoints} (CVPR2023) & 77.7 & 32.7 & 67.0 & 71.5 & 30.3 & 51.9 & 61.8 & 37.0 & 44.1 & 77.9 & 6.7 & 74.5 \\
    PLOP + \textbf{AWF(ours)} & 78.4 & 31.5 & \textbf{67.2} & 71.5 & 32.7 & \textbf{53.0} & 59.7 & 39.4 & 45.2 & 78.4 & 4.7 & 74.9 \\
    \bottomrule
  \end{tabular}
  }
  
  \caption{The mIoU(\%) of the final step on the Pascal VOC 2012 dataset for various overlapped class-incremental segmentation settings.}
  \label{tab:1}
\end{table*}
\begin{figure*}
  \centering
  \begin{subfigure}{0.49\linewidth}
     \includegraphics[height=2.3in, width=\linewidth]{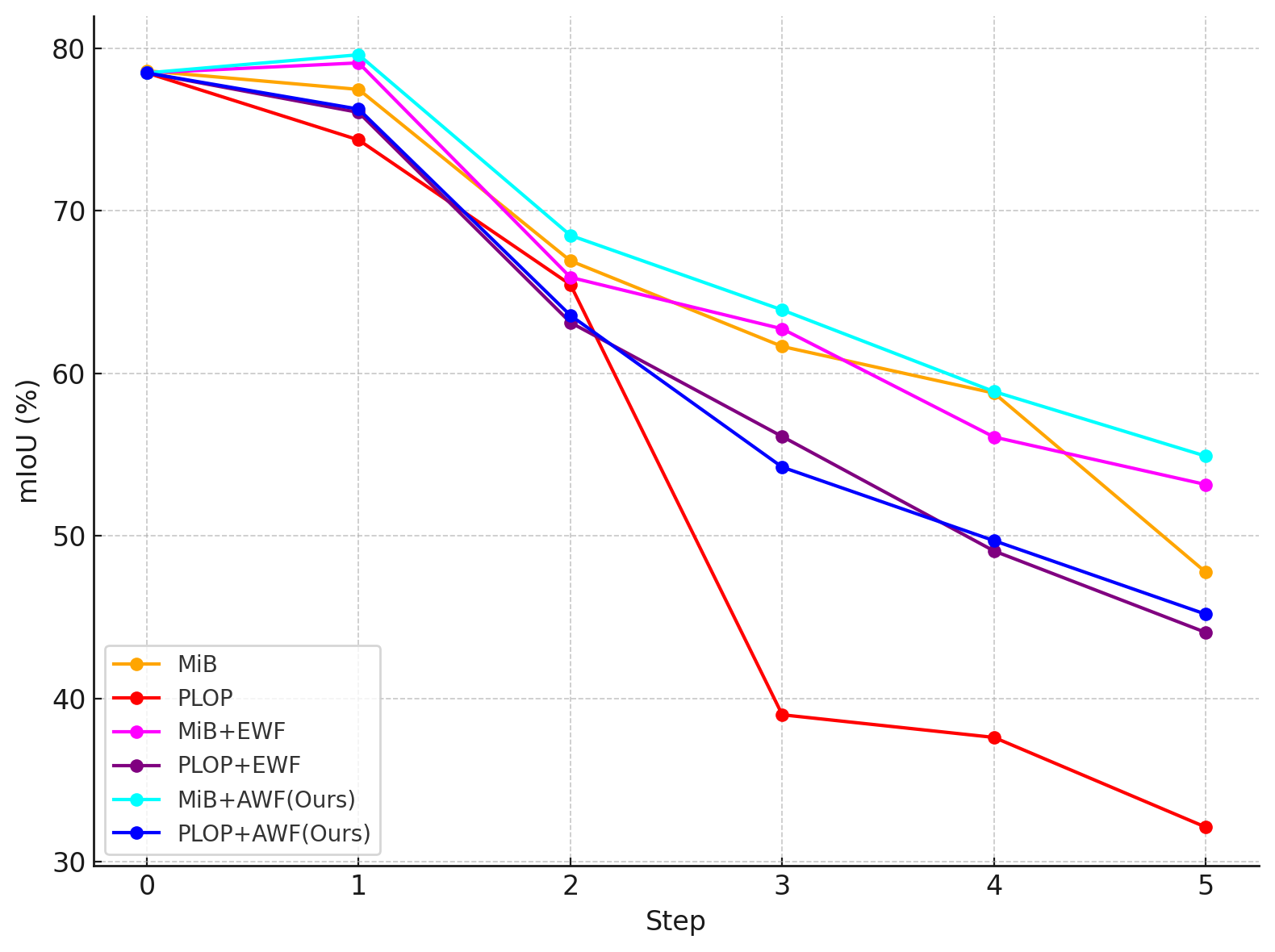}
    \caption{5-3 scenario on PASCAL VOC}
    \label{fig:3a}
  \end{subfigure}
  \hfill
  \begin{subfigure}{0.49\linewidth}
\includegraphics[height=2.3in,width=\linewidth]{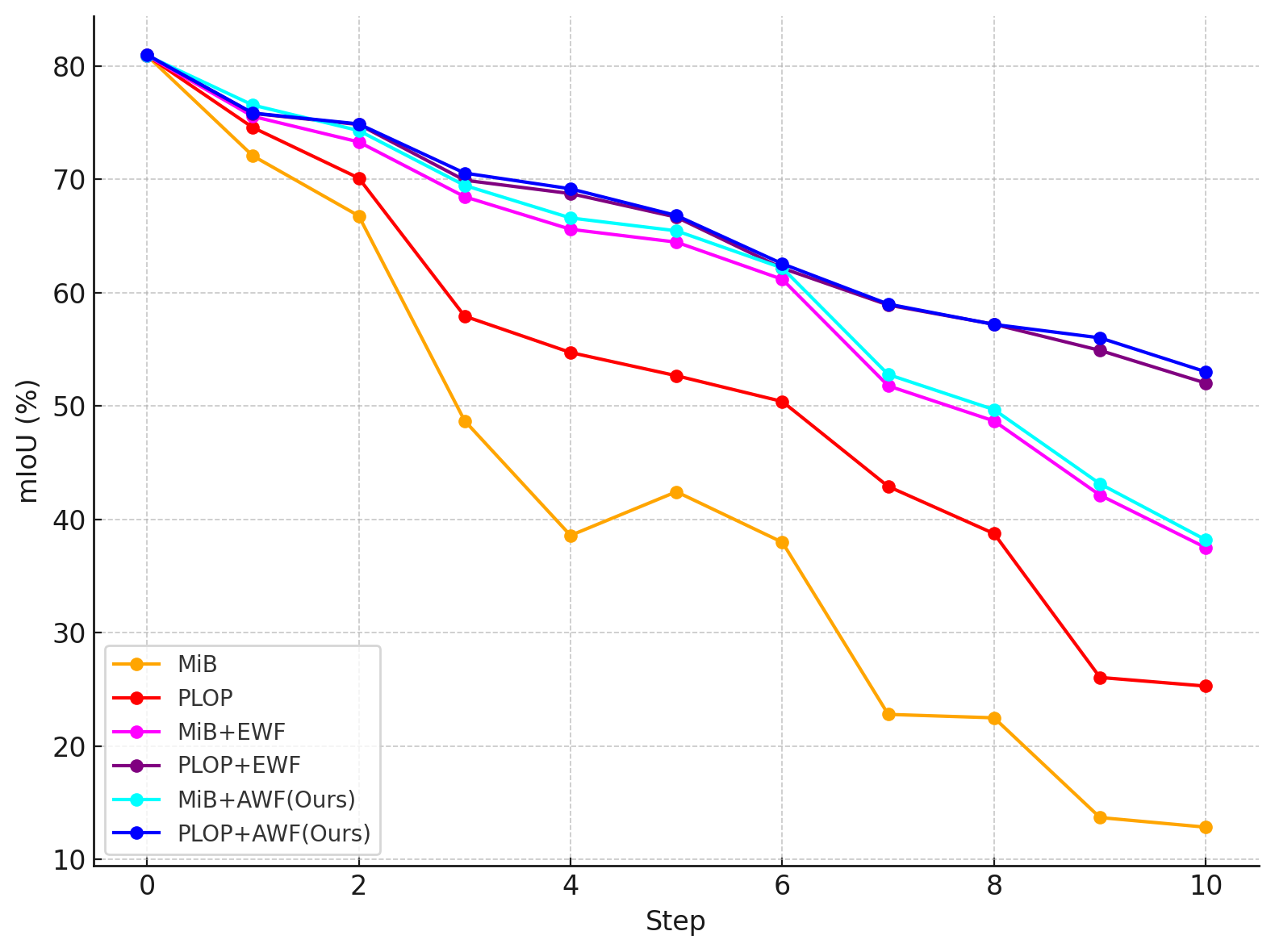}
    \caption{10-1 scenario on PASCAL VOC}
    \label{fig:3b}
  \end{subfigure}
  \caption{The mIoU (\%) at each step for the settings 5-3 (a) and 10-1 (b).}
  \label{fig:3}
\end{figure*}
\subsection{Experimental setups}
\textbf{Protocols}. In the context of Class Incremental Semantic Segmentation (CISS), the training process is generally divided into $T$ steps, where each step corresponds to a task, and the labeled classes in each task are mutually exclusive. We adopt the overlapping setting, similar to prior works, in which the current training data may contain instances that were previously labeled as background. This setup more accurately reflects real-world scenarios, and as such, we only evaluate under these conditions, consistent with previous researches \cite{NEURIPS2021_5a9542c7, Douillard_2021_CVPR}. Our experiments are conducted on two well-established segmentation datasets: PASCAL VOC 2012 \cite{pascal-voc-2012} and ADE20K \cite{Zhou_2017_CVPR}, following the existing works\cite{Douillard_2021_CVPR, Cermelli_2020_CVPR, zhang2022representation,xiao2023endpoints}.The PASCAL VOC 2012 dataset \cite{pascal-voc-2012} consists of 10,582 training images and 1,449 validation images across 20 object categories, in addition to a background class. The ADE20K dataset \cite{Zhou_2017_CVPR} comprises 150 object categories, with 20,210 images for training and 2,000 for validation. For CISS, we use the standard $A-B$ settings from previous works \cite{Douillard_2021_CVPR, Cermelli_2020_CVPR, zhang2022representation,xiao2023endpoints}, where $A$ denotes the number of classes in the initial step, and $B$ represents the number of new classes introduced in each subsequent step. At each step, only the data from the current task is available for training. On the PASCAL VOC 2012 dataset \cite{pascal-voc-2012}, we evaluate our approach with four configurations: 15-1, 10-1, 5-3, and 19-1. For the ADE20K dataset \cite{Zhou_2017_CVPR}, we test the effectiveness of our method on three configurations: 100-5, 100-10, and 100-50.
\textbf{Implementation Details}, We employ Deeplab-v3 \cite{chen2014semantic} as the segmentation model, utilizing ResNet-101 \cite{He_2016_CVPR} as the backbone, same with previous works \cite{Douillard_2021_CVPR, Cermelli_2020_CVPR, zhang2022representation,xiao2023endpoints}. For batch normalization in the backbone, we use in-place activated batch normalization \cite{bulo2018place}. We apply data augmentation techniques such as horizontal flipping and random cropping to improve model generalization. For the PASCAL VOC 2012 dataset \cite{pascal-voc-2012}, we implement an alternating training strategy, where the model is trained for a total of 45 epochs, with 30 epochs dedicated to model training and 15 epochs for optimizing the fusion parameter \(\alpha\). The alternating schedule is set with \(E_{\theta} = 10\) epochs for the model, followed by \(E_{\alpha} = 5\) epochs for \(\alpha\) optimization. For ADE20K \cite{Zhou_2017_CVPR}, the total number of training epochs is 75, with \(E_{\theta} = 20\) epochs for the model, followed by \(E_{\alpha} = 5\) epochs for \(\alpha\). The learning rate for \(\alpha\) optimization is set to \(5 \times 10^{-6}\). The training is carried out on NVIDIA A6000 GPUs with a batch size of 24. The initial learning rate for the first task is 0.01, which is reduced to 0.001 for subsequent continual learning tasks. The learning rate is decayed using a poly schedule. During training, 20\% of the training set is used for validation, We use mean Intersection over Union (mIoU) as the evaluation metric for assessing model performance.
\begin{figure*}[t]
    \centering
    \includegraphics[height=4in,width=0.97\textwidth]{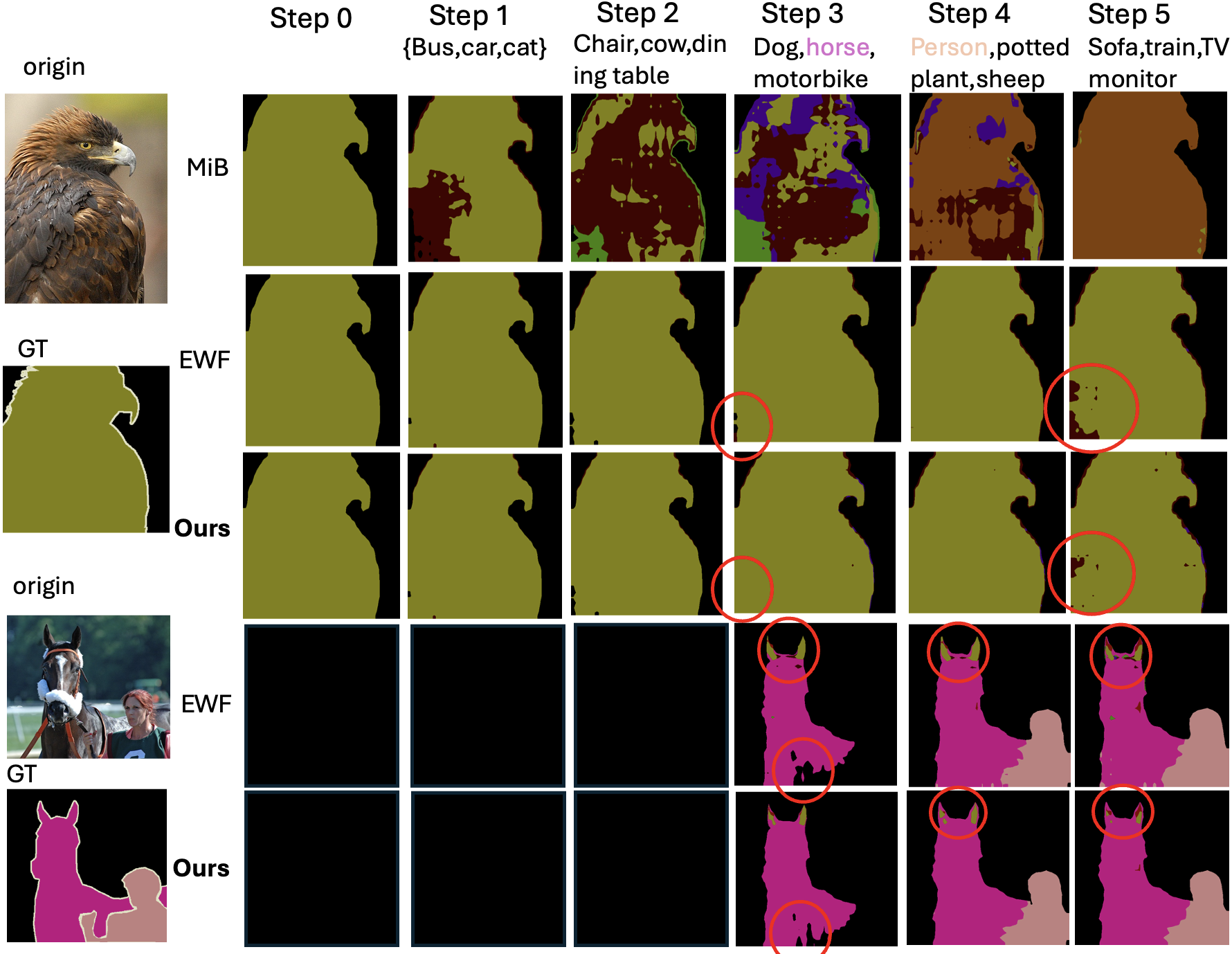}
    \caption{The qualitative comparison between different CL methods. All prediction results are from the 5-3 overlapped setting, where 5 classes are learned in step 0, followed by incremental learning of 3 classes per task in 5 subsequent tasks.}
    \label{fig:4}
\end{figure*}
\begin{table*}[t] 
  \centering
  \resizebox{\textwidth}{!}{
  {\small{
  \begin{tabular}{@{}l|ccc|ccccccc|ccc|@{}}
    \toprule
     & \multicolumn{3}{c}{\textbf{100-50} (2 steps)} & \multicolumn{7}{c}{\textbf{100-10} (6 steps)} & \multicolumn{3}{c}{\textbf{100-5} (11 steps)} \\

   \textbf{Method} & 1-100 & 101-150 & all & 1-100 & 101-110 & 111-120 & 121-130 & 131-140 & 141-150 & all & 1-100 & 101-150 & all \\
    \midrule
    ILT\cite{Michieli_2019_ICCV}  (ICCVW2019) & 18.3 & 14.8 & 17.0 & 0.1 & 0.0 & 0.1 & 0.9 & 4.1 & 9.3 & 1.1 & 0.1 & 1.3 & 0.5 \\
    PLOP\cite{Douillard_2021_CVPR} (CVPR2021) & 41.9 & 14.9 & 32.9 & 40.6 & 15.2 & 16.9 & 18.7 & 11.9 & 7.9 & 31.6 & 39.1 & 7.8 & 28.7 \\
    RC-IL\cite{zhang2022representation} (CVPR2022) & 42.3 & 18.8 & 34.5 & 39.3 & 14.6 & 26.3 & 23.2 & 12.1 & 11.8 & 32.1 & 38.5 & 11.5 & 29.6 \\
    \midrule
    MiB\cite{Cermelli_2020_CVPR} (CVPR2020) & 40.7 & 17.7 & 32.8 & 38.3 & 12.6 & 10.6 & 8.7 & 9.5 & 15.1 & 29.2 & 36.0 & 5.6 & 25.9 \\
    MiB+EWF\cite{xiao2023endpoints} (CVPR2023) & 41.2 & 21.3 & 34.6 & 41.5 & 12.8 & 22.5 & 23.2 & 14.4 & 8.8 & 33.2 & 41.4 & 13.4 & 32.1 \\
    MiB+\textbf{AWF (Ours)} & 41.8 & 23.6 & \textbf{35.8} & 42.6 & 14.3 & 25.4 & 23.7 & 14.9 & 10.1 & \textbf{34.3} & 42.4 & 16.3 & \textbf{33.8}\\
    \bottomrule
  \end{tabular}
  }
  }}
  \caption{ The mIoU(\%) of the final step on the ADE20K dataset for various overlapped continual learning settings.}
  \label{tab:2}
\end{table*}
\subsection{Comparison to baseline methods}
In this part, same as EWF\cite{xiao2023endpoints}, we apply our method to PLOP\cite{Douillard_2021_CVPR} and MiB\cite{Cermelli_2020_CVPR}. We conducted experiments on the \textbf{Pascal VOC 2012} dataset with class-incremental learning settings of 15-1, 10-1, 5-3, and 19-1. In these experiments, we compare our proposed AWF method against the EWF\cite{xiao2023endpoints}, RCIL\cite{zhang2022representation}, PLOP\cite{Douillard_2021_CVPR}, MiB\cite{Cermelli_2020_CVPR}, LwF\cite{8107520}, ILT\cite{Michieli_2019_ICCV}, and SDR\cite{Michieli_2021_CVPR}, with the results shown in Table. \ref{tab:1}. Across all settings, AWF consistently improves performance, particularly in tasks like 5-3, where more classes are added in each incremental step. In the 5-3 setting, AWF significantly outperforms MiB + EWF, improving the overall mIoU by \textbf{3.1\%}. The 5-3 setting introduces more larger number of classes at once during each incremental step, making it more challenging for EWF's alpha initialization method to adapt. In contrast, AWF's dynamic optimization of the alpha parameter leads to better performance for both old and new added classes. Similarly, in the 19-1 setting, AWF outperforms MiB+EWF and PLOP + EWF, improves the overall mIoU by \textbf{1.1\%} and \textbf{0.4\%} respectively, note that the drop in the new class performance is due to AWF, which prioritizes previously learned classes to retain knowledge. While EWF also uses a dynamic alpha, it is less effective than our adaptive approach, which better balances overall performance by focusing on older classes while slightly compromising new class accuracy because just one new classes added in last step and have 19 previous learned classes . In some typical single-step increment settings like 15-1 and 10-1, EWF's alpha initialization(Eq.\ref{eq:1}) performs well, as its alpha fusion model provides good results. However, AWF still manages to fine-tune and optimize the model performance further. For example, in the 10-1 setting, AWF improves MiB + EWF's performance by \textbf{0.9\%}. We apply our method to PLOP\cite{Douillard_2021_CVPR} which achieves a \textbf{2.4\%} improvement on the average mIoU of last ten classes while maintaining the performance on previous 10 classes at \textbf{71.5\%}, and a total \textbf{1.1\%} improvement over PLOP + EWF for average mIoU of all classes. These gains indicate that AWF enhances EWF method. In the 15-1 setting, where EWF already performs strongly, AWF maintains performance at similar levels, achieving a slight gain of \textbf{0.1\%} for MiB + EWF and \textbf{0.2\%} for PLOP + EWF. This shows that AWF can refine performance without significant deviation in tasks where EWF already achieves near-optimal results.
Overall, our AWF method dynamically optimizes the alpha during alpha training phase, particularly excelling in tasks like 5-3, where a large number of classes are added in each step. AWF also fine-tunes alpha for tasks like 15-1, preserving EWF's strong baseline performance while offering further improvements where possible. In Fig. \ref{fig:3}, we present the evolving performance across steps during the continual learning process.  As the learning progresses, our method demonstrates an increasing advantage over the baseline approaches\cite{xiao2023endpoints}.  Both MiB+AWF and PLOP+AWF exhibit strong results, maintaining higher mIoU values through the majority of the learning steps.
\textbf{Visualization}. As shown in Fig.\ref{fig:4}, we compare our method with EWF\cite{xiao2023endpoints} and MiB\cite{Cermelli_2020_CVPR}. In the third and fifth row, our method largely preserves the bird  and horse segmentation across steps, while MiB shows significant forgetting as more steps are added. In step 1 to 5 where new classes are introduced , our method successfully remain more details(\textcolor{red}{red circle}) for both old and new classes, whereas EWF begins to retain less detail in some regions. This demonstrates our method's robustness in mitigating catastrophic forgetting compared to the baselines.
knowledge better while incorporating new knowledge.
\begin{table}[h]
    \centering
    \small
    \begin{tabular}{@{}c|cccccc@{}}
        \toprule
        strategy & step1 & step2 & step3 & step4 & step5\\
        \midrule
         w/o \(\alpha\) training & 76.6 & 63.9 & 60.5 & 53.7 & 49.3 \\
        \midrule
         with \(\alpha\) training& 79.6 & 68.5 & 63.9 & 58.9 & 54.9 \\
        \bottomrule
    \end{tabular}
    \caption{Ablation study of Adaptive alpha. All performances are
reported on PASCAL VOC 2012 MiB + 5-3 overlapped setting. The values in the table are measured in mIoU (\%)}
    \label{tab:tab3}
\end{table}
To further assess the effectiveness of our approach, we performed experiments on the \textbf{ADE20K} dataset. Table.\ref{tab:2} presents the results for the 100-50, 100-10, and 100-5 settings. As seen in the table, our method demonstrates excellent performance, particularly in the more challenging 100-5 scenarios, where it achieves improvements of \textbf{7.9\%} and \textbf{5.1\%} over MiB\cite{Cermelli_2020_CVPR} and PLOP\cite{Douillard_2021_CVPR}, respectively. Additionally, our method significantly outperforms our baseline method on the 100-5 setting. where it achieves improvements of \textbf{1.7\%} over EWF\cite{xiao2023endpoints}. Compare to EWF, these results highlight the capability of our AWF approach in handling large-scale datasets more effectively.

\subsection{Ablation Study}
In this section, we demonstrate and analyze the effectiveness of our adaptive alpha training, we apply our method on MiB\cite{Cermelli_2020_CVPR}    in 5-3 task for ablation experiments.
The first row in Table.\ref{tab:tab3} represents a experiment where the model is trained for a total of 30 epochs without training the alpha. In this setup, every 10 epochs, we use alpha to fuse parameters between old model and new model. After each fusion, the model continues training for another 10 epochs and then do alpha fusion again, and this process is repeated until the 30 epochs are completed.
\textbf{In contrast}, the second row depicts the results when alpha training is incorporated. In this experiment, the model is trained for a total of 45 epochs. After each set of 10 epochs of model training, we switch to alpha training for 5 epochs. This alternating process between model and alpha training continues until all epochs are finished, allowing the alpha parameter to adapt and balance the fusion between new and old knowledge more effectively. From Table. \ref{tab:tab3}, we can see that our AWF method's \(\alpha\) optimization is effective, consistently outperforming the method without \(\alpha\) training at each step.
\begin{table}[h]
    \centering
    \small
    \begin{tabular}{@{}c|cccccc@{}}
        \toprule
        Parameter Selection & Ours & EWF & 0.2 & 0.4 & 0.6 & 0.8 \\
        \midrule
        15-1 & 65.6 & 65.6 & 65.6 & 63.7 & 60.1 & 53.3 \\
        \midrule
        10-1 & 38.2 & 37.3 & 39.5 & 31.8 & 22.7 & 14.3 \\
        \midrule
        5-3 & 54.9 & 51.8 & 38.0 & 51.2 & 56.1 & 52.9 \\
        \midrule
        Average & \textbf{52.9} & 51.6 & 47.7 & 48.9 & 46.3 & 40.2 \\
        \bottomrule
    \end{tabular}
    \caption{Comparison between AWF fusion strategy with EWF and fixed balance factors. The values in the table are measured in mIoU (\%)}
    \label{tab:4}
\end{table}

Table.\ref{tab:4} shows a comparison between our adaptive parameter fusion strategy and the EWF\cite{xiao2023endpoints} method, as well as fixed balance factors. Our method achieves the highest average performance across different settings, demonstrating its superiority over both EWF and fixed balance approaches.
\section{Conclusions}
In this paper, we introduced an Adaptive Weight Fusion (AWF) strategy aimed at improving Class Incremental Semantic Segmentation (CISS) tasks. Our AWF method leverages a dynamic, trainable fusion parameter optimized through alternating training phases, effectively balancing the retention of previously learned knowledge with the acquisition of new classes. Extensive experiments on benchmark datasets such as Pascal VOC and ADE20K demonstrate that AWF consistently outperforms the baseline Endpoints Weight Fusion (EWF)\cite{xiao2023endpoints} method and some typical methods\cite{Douillard_2021_CVPR,Cermelli_2020_CVPR} in CL field. This shows that AWF is a robust solution for mitigating catastrophic forgetting in large-scale incremental learning tasks. In the future, we plan to explore further optimizations for the fusion strategy and investigate its applicability to other continual learning domains.

\bibliographystyle{ieee_fullname}
{\small
\bibliography{references}

\begin{thebibliography}{10}\itemsep=-1pt

\bibitem{bang2021rainbow}
Jihwan Bang, Heesu Kim, YoungJoon Yoo, Jung-Woo Ha, and Jonghyun Choi.
\newblock Rainbow memory: Continual learning with a memory of diverse samples.
\newblock In {\em Proceedings of the IEEE/CVF conference on computer vision and
  pattern recognition}, pages 8218--8227, 2021.

\bibitem{Belouadah_2019_ICCV}
Eden Belouadah and Adrian Popescu.
\newblock Il2m: Class incremental learning with dual memory.
\newblock In {\em Proceedings of the IEEE/CVF International Conference on
  Computer Vision (ICCV)}, October 2019.

\bibitem{bulo2018place}
Samuel~Rota Bulo, Lorenzo Porzi, and Peter Kontschieder.
\newblock In-place activated batchnorm for memory-optimized training of dnns.
\newblock In {\em Proceedings of the IEEE conference on computer vision and
  pattern recognition}, pages 5639--5647, 2018.

\bibitem{Cermelli_2020_CVPR}
Fabio Cermelli, Massimiliano Mancini, Samuel~Rota Bulo, Elisa Ricci, and
  Barbara Caputo.
\newblock Modeling the background for incremental learning in semantic
  segmentation.
\newblock In {\em Proceedings of the IEEE/CVF Conference on Computer Vision and
  Pattern Recognition (CVPR)}, June 2020.

\bibitem{NEURIPS2021_5a9542c7}
Sungmin Cha, beomyoung kim, YoungJoon Yoo, and Taesup Moon.
\newblock Ssul: Semantic segmentation with unknown label for exemplar-based
  class-incremental learning.
\newblock In M. Ranzato, A. Beygelzimer, Y. Dauphin, P.S. Liang, and J.~Wortman
  Vaughan, editors, {\em Advances in Neural Information Processing Systems},
  volume~34, pages 10919--10930. Curran Associates, Inc., 2021.

\bibitem{Chaudhry_2018_ECCV}
Arslan Chaudhry, Puneet~K. Dokania, Thalaiyasingam Ajanthan, and Philip H.~S.
  Torr.
\newblock Riemannian walk for incremental learning: Understanding forgetting
  and intransigence.
\newblock In {\em Proceedings of the European Conference on Computer Vision
  (ECCV)}, September 2018.

\bibitem{chaudhry2021using}
Arslan Chaudhry, Albert Gordo, Puneet Dokania, Philip Torr, and David
  Lopez-Paz.
\newblock Using hindsight to anchor past knowledge in continual learning.
\newblock In {\em Proceedings of the AAAI Conference on Artificial
  Intelligence}, pages 6993--7001, 2021.

\bibitem{chen2014semantic}
Liang-Chieh Chen, George Papandreou, Iasonas Kokkinos, Kevin Murphy, and Alan~L
  Yuille.
\newblock Semantic image segmentation with deep convolutional nets and fully
  connected crfs.
\newblock {\em arXiv preprint arXiv:1412.7062}, 2014.

\bibitem{de2021continual}
Matthias De~Lange, Rahaf Aljundi, Marc Masana, Sarah Parisot, Xu Jia,
  Ale{\v{s}} Leonardis, Gregory Slabaugh, and Tinne Tuytelaars.
\newblock A continual learning survey: Defying forgetting in classification
  tasks.
\newblock {\em IEEE transactions on pattern analysis and machine intelligence},
  44(7):3366--3385, 2021.

\bibitem{Dhar_2019_CVPR}
Prithviraj Dhar, Rajat~Vikram Singh, Kuan-Chuan Peng, Ziyan Wu, and Rama
  Chellappa.
\newblock Learning without memorizing.
\newblock In {\em Proceedings of the IEEE/CVF Conference on Computer Vision and
  Pattern Recognition (CVPR)}, June 2019.

\bibitem{Ding_2019_ICCV}
Xiaohan Ding, Yuchen Guo, Guiguang Ding, and Jungong Han.
\newblock Acnet: Strengthening the kernel skeletons for powerful cnn via
  asymmetric convolution blocks.
\newblock In {\em Proceedings of the IEEE/CVF International Conference on
  Computer Vision (ICCV)}, October 2019.

\bibitem{Ding_2021_CVPR}
Xiaohan Ding, Xiangyu Zhang, Ningning Ma, Jungong Han, Guiguang Ding, and Jian
  Sun.
\newblock Repvgg: Making vgg-style convnets great again.
\newblock In {\em Proceedings of the IEEE/CVF Conference on Computer Vision and
  Pattern Recognition (CVPR)}, pages 13733--13742, June 2021.

\bibitem{Douillard_2021_CVPR}
Arthur Douillard, Yifu Chen, Arnaud Dapogny, and Matthieu Cord.
\newblock Plop: Learning without forgetting for continual semantic
  segmentation.
\newblock In {\em Proceedings of the IEEE/CVF Conference on Computer Vision and
  Pattern Recognition (CVPR)}, pages 4040--4050, June 2021.

\bibitem{douillard2020podnet}
Arthur Douillard, Matthieu Cord, Charles Ollion, Thomas Robert, and Eduardo
  Valle.
\newblock Podnet: Pooled outputs distillation for small-tasks incremental
  learning.
\newblock In {\em Computer vision--ECCV 2020: 16th European conference,
  Glasgow, UK, August 23--28, 2020, proceedings, part XX 16}, pages 86--102.
  Springer, 2020.

\bibitem{pascal-voc-2012}
M. Everingham, L. Van~Gool, C.~K.~I. Williams, J. Winn, and A. Zisserman.
\newblock The {PASCAL} {V}isual {O}bject {C}lasses {C}hallenge 2012 {(VOC2012)}
  {R}esults.
\newblock
  http://www.pascal-network.org/challenges/VOC/voc2012/workshop/index.html.

\bibitem{garcia2017review}
Alberto Garcia-Garcia, Sergio Orts-Escolano, Sergiu Oprea, Victor
  Villena-Martinez, and Jose Garcia-Rodriguez.
\newblock A review on deep learning techniques applied to semantic
  segmentation.
\newblock {\em arXiv preprint arXiv:1704.06857}, 2017.

\bibitem{grigorescu2020survey}
Sorin Grigorescu, Bogdan Trasnea, Tiberiu Cocias, and Gigel Macesanu.
\newblock A survey of deep learning techniques for autonomous driving.
\newblock {\em Journal of field robotics}, 37(3):362--386, 2020.

\bibitem{he2019rethinking}
Kaiming He, Ross Girshick, and Piotr Doll{\'a}r.
\newblock Rethinking imagenet pre-training.
\newblock In {\em Proceedings of the IEEE/CVF international conference on
  computer vision}, pages 4918--4927, 2019.

\bibitem{He_2016_CVPR}
Kaiming He, Xiangyu Zhang, Shaoqing Ren, and Jian Sun.
\newblock Deep residual learning for image recognition.
\newblock In {\em Proceedings of the IEEE Conference on Computer Vision and
  Pattern Recognition (CVPR)}, June 2016.

\bibitem{jiang2018medical}
Feng Jiang, Aleksei Grigorev, Seungmin Rho, Zhihong Tian, YunSheng Fu, Worku
  Jifara, Khan Adil, and Shaohui Liu.
\newblock Medical image semantic segmentation based on deep learning.
\newblock {\em Neural Computing and Applications}, 29:1257--1265, 2018.

\bibitem{Kim_2021_ICCV}
Chris~Dongjoo Kim, Jinseo Jeong, Sangwoo Moon, and Gunhee Kim.
\newblock Continual learning on noisy data streams via self-purified replay.
\newblock In {\em Proceedings of the IEEE/CVF International Conference on
  Computer Vision (ICCV)}, pages 537--547, October 2021.

\bibitem{Kornblith_2019_CVPR}
Simon Kornblith, Jonathon Shlens, and Quoc~V. Le.
\newblock Do better imagenet models transfer better?
\newblock In {\em Proceedings of the IEEE/CVF Conference on Computer Vision and
  Pattern Recognition (CVPR)}, June 2019.

\bibitem{8107520}
Zhizhong Li and Derek Hoiem.
\newblock Learning without forgetting.
\newblock {\em IEEE Transactions on Pattern Analysis and Machine Intelligence},
  40(12):2935--2947, 2018.

\bibitem{Liu_2021_CVPR}
Yaoyao Liu, Bernt Schiele, and Qianru Sun.
\newblock Adaptive aggregation networks for class-incremental learning.
\newblock In {\em Proceedings of the IEEE/CVF Conference on Computer Vision and
  Pattern Recognition (CVPR)}, pages 2544--2553, June 2021.

\bibitem{mccloskey1989catastrophic}
Michael McCloskey and Neal~J Cohen.
\newblock Catastrophic interference in connectionist networks: The sequential
  learning problem.
\newblock In {\em Psychology of learning and motivation}, volume~24, pages
  109--165. Elsevier, 1989.

\bibitem{Michieli_2019_ICCV}
Umberto Michieli and Pietro Zanuttigh.
\newblock Incremental learning techniques for semantic segmentation.
\newblock In {\em Proceedings of the IEEE/CVF International Conference on
  Computer Vision (ICCV) Workshops}, Oct 2019.

\bibitem{michieli2021continual}
Umberto Michieli and Pietro Zanuttigh.
\newblock Continual semantic segmentation via repulsion-attraction of sparse
  and disentangled latent representations.
\newblock In {\em Proceedings of the IEEE/CVF conference on computer vision and
  pattern recognition}, pages 1114--1124, 2021.

\bibitem{Michieli_2021_CVPR}
Umberto Michieli and Pietro Zanuttigh.
\newblock Continual semantic segmentation via repulsion-attraction of sparse
  and disentangled latent representations.
\newblock In {\em Proceedings of the IEEE/CVF Conference on Computer Vision and
  Pattern Recognition (CVPR)}, pages 1114--1124, June 2021.

\bibitem{pathak2018application}
Ajeet~Ram Pathak, Manjusha Pandey, and Siddharth Rautaray.
\newblock Application of deep learning for object detection.
\newblock {\em Procedia computer science}, 132:1706--1717, 2018.

\bibitem{article}
Boris Polyak and Anatoli Juditsky.
\newblock Acceleration of stochastic approximation by averaging.
\newblock {\em SIAM Journal on Control and Optimization}, 30:838--855, 07 1992.

\bibitem{Rebuffi_2017_CVPR}
Sylvestre-Alvise Rebuffi, Alexander Kolesnikov, Georg Sperl, and Christoph~H.
  Lampert.
\newblock icarl: Incremental classifier and representation learning.
\newblock In {\em Proceedings of the IEEE Conference on Computer Vision and
  Pattern Recognition (CVPR)}, July 2017.

\bibitem{Simon_2021_CVPR}
Christian Simon, Piotr Koniusz, and Mehrtash Harandi.
\newblock On learning the geodesic path for incremental learning.
\newblock In {\em Proceedings of the IEEE/CVF Conference on Computer Vision and
  Pattern Recognition (CVPR)}, pages 1591--1600, June 2021.

\bibitem{singh2021rectification}
Pravendra Singh, Pratik Mazumder, Piyush Rai, and Vinay~P Namboodiri.
\newblock Rectification-based knowledge retention for continual learning.
\newblock In {\em Proceedings of the IEEE/CVF conference on computer vision and
  pattern recognition}, pages 15282--15291, 2021.

\bibitem{NEURIPS2020_b3b43aee}
Pravendra Singh, Vinay~Kumar Verma, Pratik Mazumder, Lawrence Carin, and Piyush
  Rai.
\newblock Calibrating cnns for lifelong learning.
\newblock In H. Larochelle, M. Ranzato, R. Hadsell, M.F. Balcan, and H. Lin,
  editors, {\em Advances in Neural Information Processing Systems}, volume~33,
  pages 15579--15590. Curran Associates, Inc., 2020.

\bibitem{Smith_2021_ICCV}
James Smith, Yen-Chang Hsu, Jonathan Balloch, Yilin Shen, Hongxia Jin, and
  Zsolt Kira.
\newblock Always be dreaming: A new approach for data-free class-incremental
  learning.
\newblock In {\em Proceedings of the IEEE/CVF International Conference on
  Computer Vision (ICCV)}, pages 9374--9384, October 2021.

\bibitem{Verma_2021_CVPR}
Vinay~Kumar Verma, Kevin~J Liang, Nikhil Mehta, Piyush Rai, and Lawrence Carin.
\newblock Efficient feature transformations for discriminative and generative
  continual learning.
\newblock In {\em Proceedings of the IEEE/CVF Conference on Computer Vision and
  Pattern Recognition (CVPR)}, pages 13865--13875, June 2021.

\bibitem{wang2022foster}
Fu-Yun Wang, Da-Wei Zhou, Han-Jia Ye, and De-Chuan Zhan.
\newblock Foster: Feature boosting and compression for class-incremental
  learning.
\newblock In {\em European conference on computer vision}, pages 398--414.
  Springer, 2022.

\bibitem{xiao2023endpoints}
Jia-Wen Xiao, Chang-Bin Zhang, Jiekang Feng, Xialei Liu, Joost van~de Weijer,
  and Ming-Ming Cheng.
\newblock Endpoints weight fusion for class incremental semantic segmentation.
\newblock In {\em Proceedings of the IEEE/CVF Conference on Computer Vision and
  Pattern Recognition}, pages 7204--7213, 2023.

\bibitem{Yan_2021_CVPR}
Shipeng Yan, Jiangwei Xie, and Xuming He.
\newblock Der: Dynamically expandable representation for class incremental
  learning.
\newblock In {\em Proceedings of the IEEE/CVF Conference on Computer Vision and
  Pattern Recognition (CVPR)}, pages 3014--3023, June 2021.

\bibitem{zhang2022representation}
Chang-Bin Zhang, Jia-Wen Xiao, Xialei Liu, Ying-Cong Chen, and Ming-Ming Cheng.
\newblock Representation compensation networks for continual semantic
  segmentation.
\newblock In {\em Proceedings of the IEEE/CVF Conference on Computer Vision and
  Pattern Recognition}, pages 7053--7064, 2022.

\bibitem{Zhou_2017_CVPR}
Bolei Zhou, Hang Zhao, Xavier Puig, Sanja Fidler, Adela Barriuso, and Antonio
  Torralba.
\newblock Scene parsing through ade20k dataset.
\newblock In {\em Proceedings of the IEEE Conference on Computer Vision and
  Pattern Recognition (CVPR)}, July 2017.

\bibitem{Zhu_2022_CVPR}
Kai Zhu, Wei Zhai, Yang Cao, Jiebo Luo, and Zheng-Jun Zha.
\newblock Self-sustaining representation expansion for non-exemplar
  class-incremental learning.
\newblock In {\em Proceedings of the IEEE/CVF Conference on Computer Vision and
  Pattern Recognition (CVPR)}, pages 9296--9305, June 2022.

\end{thebibliography}
}
\end{document}